\documentclass{article}

\usepackage{microtype}
\usepackage{graphicx}
\usepackage{subcaption}
\usepackage{booktabs} 

\usepackage{xurl} 
\usepackage[hypertexnames=false]{hyperref}

\usepackage[preprint]{icml2026}

\usepackage{amsmath}
\usepackage{amssymb}
\usepackage{mathtools}
\usepackage{amsthm}
\usepackage{graphicx} 

\usepackage[most]{tcolorbox}
\usepackage{xcolor}

\newtcolorbox{graynote}{
  colback=gray!12,      
  colframe=gray!55,     
  boxrule=0.6pt,        
  arc=0.5mm,              
  left=5pt,right=5pt,top=2pt,bottom=2pt, 
  width=\linewidth
}

\usepackage[capitalize,noabbrev]{cleveref}

\theoremstyle{plain}

\theoremstyle{definition}

\theoremstyle{remark}

\usepackage{booktabs}
\usepackage{multirow}
\usepackage{makecell}
\usepackage{adjustbox}   
\usepackage{xcolor}
\usepackage{colortbl}  

\usepackage[textsize=tiny]{todonotes}
\usepackage{xspace}

\icmltitlerunning{ConFoThinking: Consolidated Focused Attention Driven Thinking for Visual Question Answering}
\newcommand{\Method}{ConFoThinking\xspace}
\definecolor{darkgreen}{RGB}{0,100,0}
\begin{document}

\twocolumn[
  \icmltitle{\Method: Consolidated Focused Attention Driven Thinking for Visual Question Answering
  }



  \icmlsetsymbol{equal}{*}

  \begin{icmlauthorlist}
    \icmlauthor{Zhaodong Wu}{equal,Pku,XJTLU}
    \icmlauthor{Haochen Xue}{equal,XJTLU}
    \icmlauthor{Qi Cao}{equal,XJTLU}
    \icmlauthor{Wenqi Mo}{XJTLU}
    \icmlauthor{Yu Pei}{XJTLU}
    \icmlauthor{Wenqi Xu}{XJTLU}
    \icmlauthor{Jionglong Su}{XJTLU}
    \icmlauthor{Yang Liu}{Pku}

  \end{icmlauthorlist}

  \icmlaffiliation{Pku}{Wangxuan Institute of Computer Technology, Peking University, Beijing, China}
  \icmlaffiliation{XJTLU}{School of AI and Advanced Computing, Xi'an Jiaotong-Liverpool University, Suzhou, China}

  \icmlcorrespondingauthor{Yang Liu}{yangliu@pku.edu.cn}

  \icmlkeywords{Machine Learning, ICML}

  \vskip 0.3in
]



\printAffiliationsAndNotice{}  

\begin{figure*}[h]
  \centering
  \includegraphics[width=\textwidth]{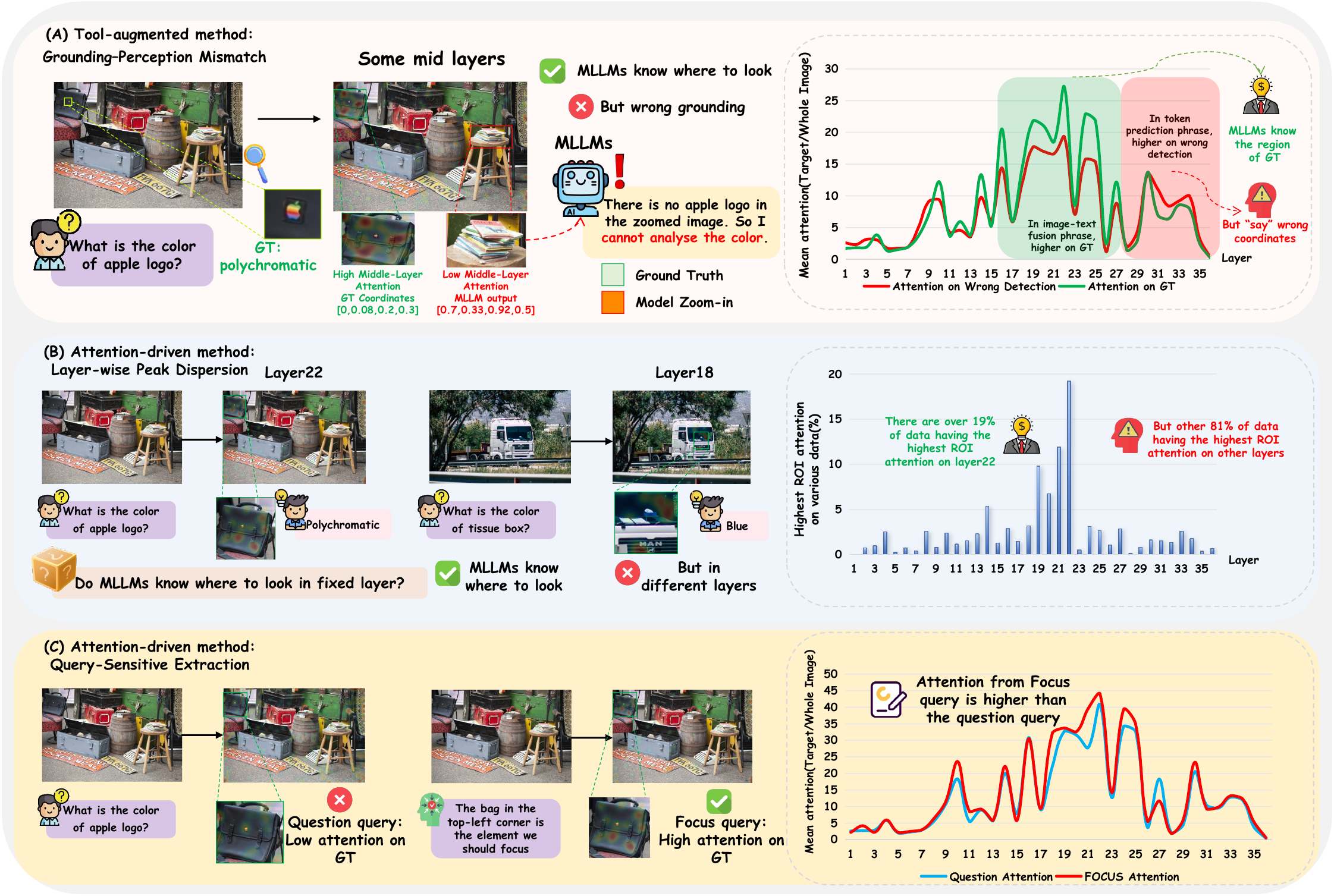}
  \caption{Limitations of existing methods. (A) In tool-augmented coordinate-output pipelines, MLLMs may ``say" incorrect bounding-box coordinates, even though their intermediate vision-language fusion layers still attend to the GT ROI. The line chart shows the attention distribution inside the predicted and GT boxes for Qwen3-VL-8B (with tools) on V* benchmark~\cite{wu2024v}, where the model outputs an incorrect answer and the predicted box has IoU $<$ 0.1 with the ground-truth box. (B) In attention-driven methods, where-to-look signals are fragmented across layers, making any fixed-layer choice unreliable. The bar chart shows the distribution of the highest-ROI-attention layer for Qwen3-VL-8B across samples in VisCoT (only 19.3\% at a fixed single layer (layer22)). (C) Where-to-look signals are query-sensitive: extracting attention from semantic visual cues is more accurate than extracting it from the raw question. The line chart compares the layer-wise attention distributions of these two approaches on the V* benchmark. 
}
  \label{fig:methods_comparison}
\end{figure*}

\begin{abstract}
Thinking with Images improves fine-grained VQA for MLLMs by emphasizing visual cues. However, tool-augmented methods depend on the capacity of grounding, which remains unreliable for MLLMs. In parallel, attention-driven methods to crop the Region of Interest (ROIs) are proposed but they are constrained by (1) fragmented attention signals scattered across layers, leading to suboptimal localization, (2) relying on question- or redundant-text-conditioned attention extraction. Our analysis reveals three patterns: MLLMs may attend to the correct region yet generate incorrect coordinates, where-to-look attention is often fragmented across layers, and attention extraction is query-sensitive. Motivated by these, We propose \textbf{\Method}, a \textbf{Con}solidated-\textbf{Fo}cused-Attention-Driven \textbf{Thinking} framework that learns to aggregate attention into a designated intermediate layer, from which we mine and zoom in salient regions for downstream visual understanding. Moreover, we extract attention using concise semantic cues of what to look into, which mitigates the semantic noise introduced by question- or redundant-text-based attention extraction. Experiments across five VQA benchmarks demonstrate \textbf{\Method} significantly improves perception performance. The code, checkpoints, and dataset will be released after being accepted.

\end{abstract}
\section{Introduction}

Multimodal Large Language Models (MLLMs) have made rapid progress on vision--language understanding and are increasingly used for tasks that require step-by-step reasoning~\cite{wei2022chain}. In fine-grained VQA (especially on high-resolution images), errors often come from missing the right visual evidence rather than lacking reasoning ability. Therefore, many recent ``Thinking with Images'' pipelines introduce image-space operations such as cropping and zooming to acquire additional evidence. Crucially, these operations are only effective if the model can reliably decide \emph{where to look} first.

ROI localization is thus the key interface that connects ``thinking'' to ``seeing'' in Thinking with Images. Existing methods typically follow two routes. The first route asks the model to directly output bounding-box coordinates to drive cropping/zooming~\cite{wang2025pixel}. While convenient, it is fragile: bounding-box coordinates are continuous geometric variables, yet the model must emit them as discrete tokens (typically formatted numbers) under an autoregressive next-token objective. Consequently, the decoded box can be inaccurate even when the relevant visual evidence has been internally localized (Fig.~\ref{fig:methods_comparison}(a)). The second route extracts ROIs from attention maps, such as ICoT~\cite{gao2025interleaved}, ViCrop~\cite{zhang2025mllms}, and Focus~\cite{zhong2025focus}. This avoids explicit coordinate generation, but it introduces two practical issues: (i) attention is task- and layer-dependent~\cite{shi2025vision}, making a fixed-layer choice fragile (Fig.~\ref{fig:methods_comparison}(b)); and (ii) extracting attention using long questions or redundant text produces diffuse, noisy heatmaps, 
which undermines precise ROI mining (Fig.~\ref{fig:methods_comparison}(c)). Overall, existing pipelines lack a stable and low-noise attention-based mechanism that can robustly produce ROIs across tasks and datasets.

To understand this gap, we empirically analyze \emph{where-to-look} signals in MLLMs under the Thinking with Images paradigm. The curves in Fig~\ref{fig:methods_comparison}(a) reveal a consistent mismatch between explicit coordinate outputs and internal where-to-look signals. In failure cases with incorrect predicted boxes, text-to-image attention in intermediate vision--language fusion layers assigns higher mass to the ground-truth ROI, whereas attention in late coordinate-decoding layers becomes biased toward the predicted (but incorrect) region.
This stage-wise shift suggests that coordinate decoding can deviate from earlier cross-modal localization due to late-layer attention drift during token generation. The bar chart in Fig.~\ref{fig:methods_comparison}(b) shows that peak where-to-look attention is highly sample-dependent on VisCoT~\cite{shao2024visual}: even the most frequent peak layer (Layer~22) covers only 19.3\% of samples, and peaks are spread across many other layers. 
This dispersion indicates that extracting ROIs from a single fixed layer is far from optimal for a large fraction of inputs. The curves in Fig.~\ref{fig:methods_comparison}(c) compare two attention-extraction queries: the red curve aggregates attention from a semantically guided visual cue, while the blue curve aggregates attention from the raw question. Across most layers, the cue-conditioned attention is consistently higher than the question-conditioned attention. This suggests that constructing semantic visual cues can provide a stronger where-to-look signal to facilitate visual reasoning.

These observations lead to a simple goal:
\vspace{-5pt}
\begin{graynote}
                    \textit{Make the model's where-to-look attention reliable for cropping by consolidating it into a fixed layer, and use it as a semantic focus signal to mine ROIs.}
\end{graynote}
\vspace{-4pt}

To this end, we propose Consolidated-Focused-Attention-Driven Thinking (\Method), a new Thinking with Images framework that does \emph{not} require the MLLM to explicitly generate box coordinates. Our key idea is to disentangle \emph{what to look for} from \emph{where to look}. Specifically, ConFoAttn first generates a concise semantic focus cue wrapped by \texttt{<FOCUS>}...\texttt{</FOCUS>}, which specifies \emph{what} visual evidence should be inspected. Conditioned on this cue, we extract text-to-image attention and refine it with bounding-box supervision to condense the \emph{where-to-look} signal into a designated intermediate layer (e.g., Layer~22 for Qwen3-VL-8B, yielding a stable single-layer \emph{semantic focus map} for ROI mining (Fig.~\ref{fig:methods_comparison}(c)). Finally, we train a heatmap-to-box predictor, AttnDetector, to convert the refined focus map into bounding-box coordinates for accurate and robust zooming.

Our contributions are summarized as follows:
\vspace{-12pt}
\begin{itemize}
  \item We conduct an interpretability study of where-to-look signals in MLLMs and uncover two failure modes: (i) a grounding--perception mismatch in coordinate-output pipelines (models may attend to the GT ROI but decode incorrect boxes), and (ii) fragmented where-to-look attention across layers, which explains the instability of fixed-layer attention cropping.

  \vspace{-8pt}
  \item   We propose \Method, which disentangles \emph{what to look for} (a concise \texttt{<FOCUS>} cue for querying attention) from \emph{where to look} (box-supervised attention condensation into a designated intermediate layer), and localizes ROIs via a learned heatmap-to-box predictor (AttnDetector) for evidence-aware VQA.

  \vspace{-8pt}
  \item Extensive comparisons and ablations demonstrate that \Method achieves state-of-the-art performance on five VQA benchmarks.

\end{itemize}

\vspace{-15pt}
\section{Related Work}

\subsection{Thinking with Images via Tool-based ROI Selection}
Thinking with images extends chain-of-thought reasoning to vision language settings by enabling models to actively acquire visual evidence during reasoning~\cite{o3,liu2025reasoning,liu2025more}. A major line of work obtains regions of interest (ROIs) via explicit image-space actions. Training-free methods iteratively propose regions and perform search to identify informative ROIs without additional training~\cite{liu2024chain,shen2025zoomeye,wang2025visuothink}. Supervised fine-tuning (SFT) improves grounding and localization by supervising intermediate steps with visual evidence or directly predicting target regions~\cite{sun2024visual,zhang2025cmmcot,zhao2025pyvision,fu2025refocus}. More recently, reinforcement learning (RL) methods, often motivated by objectives such as GRPO~\cite{shao2024deepseekmath}, warm-start from SFT and further optimize exploration and region selection~\cite{zhang2025thyme,zheng2025deepeyes,hong2025deepeyesv2,wang2025pixel}. However, most of these pipelines still require the MLLM to decode pixel-space bounding-box coordinates to trigger crop actions. This step is unreliable because coordinates are continuous geometric variables but must be emitted as discrete tokens under autoregressive decoding, so small numeric errors can cause large box shifts; moreover, our analysis shows that coordinate-decoding layers can drift from earlier cross-modal localization, yielding incorrect ROIs even when the model attends to the correct region (Fig.~\ref{fig:methods_comparison}(a)).
\vspace{-5pt}
\subsection{Attention-driven ROI Cropping}
More closely related to our work are methods that avoid explicit coordinate decoding and instead derive ROIs from internal model signals. ICoT~\cite{gao2025interleaved} and ViCrop~\cite{zhang2025mllms} guide cropping by extracting attention maps from the MLLM and selecting high-attention regions for zooming. FOCUS~\cite{zhong2025focus} similarly localizes ROIs from model-internal signals to focus the visual evidence used for answering. These approaches are attractive because they use the model’s own where-to-look cues, but they are often sensitive to \emph{how} the signal is extracted: where-to-look attention can vary substantially across layers and can become diffuse when conditioned on long questions or redundant text, making ROI quality unstable across samples. In contrast, our method explicitly disentangles \emph{what to look for} (a concise \texttt{<FOCUS>} cue) from \emph{where to look} (box-supervised attention condensation into a designated intermediate layer), producing stable fixed-layer heatmaps that are further converted to boxes by a learned heatmap-to-box predictor for robust ROI mining.
\vspace{-10pt}
\section{Where-to-Look Signals in MLLMs: An Empirical Study}

In Thinking with Images pipelines, overall performance often hinges on whether the ROI is localized correctly. In this section, we empirically analyze where-to-look signals in MLLMs to diagnose two major failure modes (tool-augmented method and attention-driven method) in ROI localization and motivate our attention-refinement design. We focus on three research questions (RQs):
\vspace{-0.35cm}
\begin{itemize}
    \item RQ1: When MLLMs output incorrect ROI coordinates, does their internal attention still favor the ground-truth target region?
    \item RQ2: Is the layer that exhibits the strongest where-to-look attention consistent across samples?
    \item RQ3: When extracting text-to-image attention for ROI mining, which text query is more reliable?
\end{itemize}

\vspace{-10pt}
\subsection{Grounding--Perception Mismatch in Coordinate Output}\label{section3.1}
We first study coordinate-output pipelines. Let $\mathcal{R}_{\text{pred}}$ denote the set of tokens in ROI box produced by coordinate generation and $\mathcal{R}_{\text{GT}}$ those in ground-truth target ROI. We call a sample a \emph{grounding-error} case when $\mathrm{IoU}(\mathcal{R}_{\text{pred}},\mathcal{R}_{\text{GT}})<\tau$ (we use $\tau{=}0.1$). We use \emph{perception} to refer to the model's internal where-to-look attention, and test the hypothesis:
\begin{graynote}
\textit{In grounding-error cases, if coordinate outputs faithfully reflect the model's perception, then the model's attention should align more with $\mathcal{R}_{\text{pred}}$ than with $\mathcal{R}_{\text{GT}}$.}
\end{graynote}
To evaluate the hypothesis, we run the same Thinking with Images setup on V* Bench with two representative coordinate-output pipelines: Qwen3-VL equipped with a zoom-in tool and Pixel Reasoner. For each pipeline, we collect grounding-error cases where the final answer is wrong and compare attention concentration on $\mathcal{R}_{\text{GT}}$ versus $\mathcal{R}_{\text{pred}}$.

To quantify attention concentration on an ROI, for any region $\mathcal{R}\subseteq\mathcal{R}_{\text{img}}$ we define the following normalized mean-attention ratio at layer $\ell$:
\begin{equation}
s^{(\ell)}(\mathcal{R}) =
\frac{\frac{1}{|\mathcal{R}|}\sum_{p \in \mathcal{R}} \bar{a}^{(\ell)}_{p}}
{\frac{1}{|\mathcal{R}_{\text{img}}|}\sum_{p \in \mathcal{R}_{\text{img}}} \bar{a}^{(\ell)}_{p}}
\label{eq:attn_mean_ratio}
\end{equation}
For a given sample, let $\mathcal{R}_{\text{img}}$ denote the set of all image tokens/patches of that image. Any region $\mathcal{R}$ is a subset of image tokens, i.e., $\mathcal{R}\subseteq\mathcal{R}_{\text{img}}$. The score $s^{(\ell)}(\mathcal{R})$ measures how strongly layer-$\ell$ attention concentrates on region $\mathcal{R}$, normalized by the mean attention over the whole image.
\begin{equation}
\bar{a}^{(\ell)}_{p}
=
\frac{1}{|\mathcal{Q}|\,H}
\sum_{q \in \mathcal{Q}}
\sum_{h=1}^{H}
A^{(\ell,h)}_{q \rightarrow p}
\label{eq:attn_map_aggregation}
\end{equation}
And for Eq.~\ref{eq:attn_map_aggregation}, $\mathcal{Q}$ denotes the set of text tokens, such as the question and generated reasoning tokens. $H$ is the number of attention heads, and $A^{(\ell,h)}_{q \rightarrow p}$ is the attention weight from text token $q$ to image token $p$ at layer $\ell$ and head $h$. Fig.~\ref{fig:methods_comparison}(a) plots the layer-wise \emph{average} of $s^{(\ell)}(\mathcal{R}_{\text{GT}})$ (green) and $s^{(\ell)}(\mathcal{R}_{\text{pred}})$ (red) over all grounding-error cases on V* Bench for Qwen3-VL-8B.

 We observe a clear stage-wise split. In intermediate vision--language fusion layers (where cross-modal evidence is integrated and the model decides \emph{where to look}~\cite{neo2024towards}), attention concentrates more on the true target, i.e., $s^{(\ell)}(\mathcal{R}_{\text{GT}}) > s^{(\ell)}(\mathcal{R}_{\text{pred}})$. Toward late decoding layers that support token generation (including coordinate tokens), attention can drift toward the predicted but incorrect region, making $s^{(\ell)}(\mathcal{R}_{\text{pred}})$ surpass $s^{(\ell)}(\mathcal{R}_{\text{GT}})$. We observe the same stage-wise pattern for Pixel Reasoner; the corresponding plot is provided in Appendix~\ref{Appendix:Grounding-Perception_Mismatch}. These suggests a mechanistic mismatch: fusion-stage layers already capture the correct evidence for \emph{where to look}, but coordinate tokens are decoded in late layers where attention drift can dominate, causing the model to output an incorrect box despite attending to the right region.
\vspace{-5pt}
\subsection{Fragmented Where-to-Look Signals Across Layers}\label{section3.2}
Given the grounding--perception mismatch in Section~\ref{section3.1}, a natural alternative is to extract ROIs from attention. However, attention-driven cropping is often unreliable in practice. A key reason is that where-to-look attention is fragmented across layers, making ROI extraction highly sensitive to layer choice. For each sample, we define the peak-attention layer index as the layer that maximizes the attention score on the ground-truth target region:
\begin{equation}\label{eq:Peak_attention}
\ell^{\star} \;=\; \arg\max_{\ell \in \{1,\dots,L\}} \; s^{(\ell)}(\mathcal{R}_{\text{tar}}) ,
\end{equation}
As shown in Fig.~1(b) on the validation set, VisCoT (Qwen3-VL-8B), the peak-attention layer $\ell^\star$ varies substantially across samples: even the most frequent peak layer accounts for only about 19.3\% of cases, while peaks are spread across many other layers. This dispersed peak-layer pattern also holds for other backbones (e.g., Qwen3-VL-4B and Qwen2.5-VL-7B; Appendix~\ref{app:peak-layer-other-backbones}). Consequently, extracting attention from a single fixed layer is far from optimal for many inputs, and naive aggregation across layers can also blur the ROI signal.

\vspace{-5pt}
\subsection{Query-Sensitive Attention Extraction for ROI Mining}\label{section3.3}

We study RQ3: \emph{when mining ROIs from text-to-image attention, which query is more reliable---the raw question or a concise semantic cue?}
Specifically, we compute the same text-to-image attention statistic but change the \emph{query} tokens: (i) the question tokens, or (ii) the tokens generated by GPT-5 conditioned on the image and question, which are the key semantically guided visual cues.
Fig.~\ref{fig:methods_comparison}(c) plots the layer-wise \emph{average} attention concentration on the ground-truth ROI, i.e., $s^{(\ell)}(\mathcal{R}_{\text{tar}})$, on the V* test set.
Across most layers, using semantically guided visual cues as the query yields higher ROI attention concentration than using the raw question (red vs.\ blue), indicating a more ROI-specific where-to-look signal.

This result suggests that attention-based localization is \emph{query-sensitive}. The raw question may contain auxiliary or abstract phrasing that is not tied to a specific region, which can weaken ROI-specific attention. In contrast, semantically guided visual cues narrows the query to the intended visual evidence to inspect, leading to higher attention concentration on the target ROI. Therefore, we extract where-to-look heatmaps using semantically guided visual cues rather than the full question, complementing the layer-fragmentation finding in Section~\ref{section3.2} and motivating our attention-refinement design.


\vspace{-10pt}
\paragraph{Summary and Motivation.}
Taken together, our analyses show that MLLMs often contain correct internal where-to-look signals, but existing pipelines fail to reliably convert them into accurate and reusable ROIs: coordinate-output pipelines can exhibit a grounding--perception mismatch (Section~\ref{section3.1}), layer-wise attention peaks are fragmented so fixed-layer cropping is unstable (Section~\ref{section3.2}), and ROI mining is also sensitive to the text query (Section~\ref{section3.3}). Motivated by these findings, we propose \Method, which uses a concise \texttt{<FOCUS>} span to query attention and condenses where-to-look attention into a designated intermediate layer, enabling stable heatmap extraction and reliable ROI localization for downstream VQA.
\begin{figure*}[t]
  \centering
  \includegraphics[width=\textwidth]{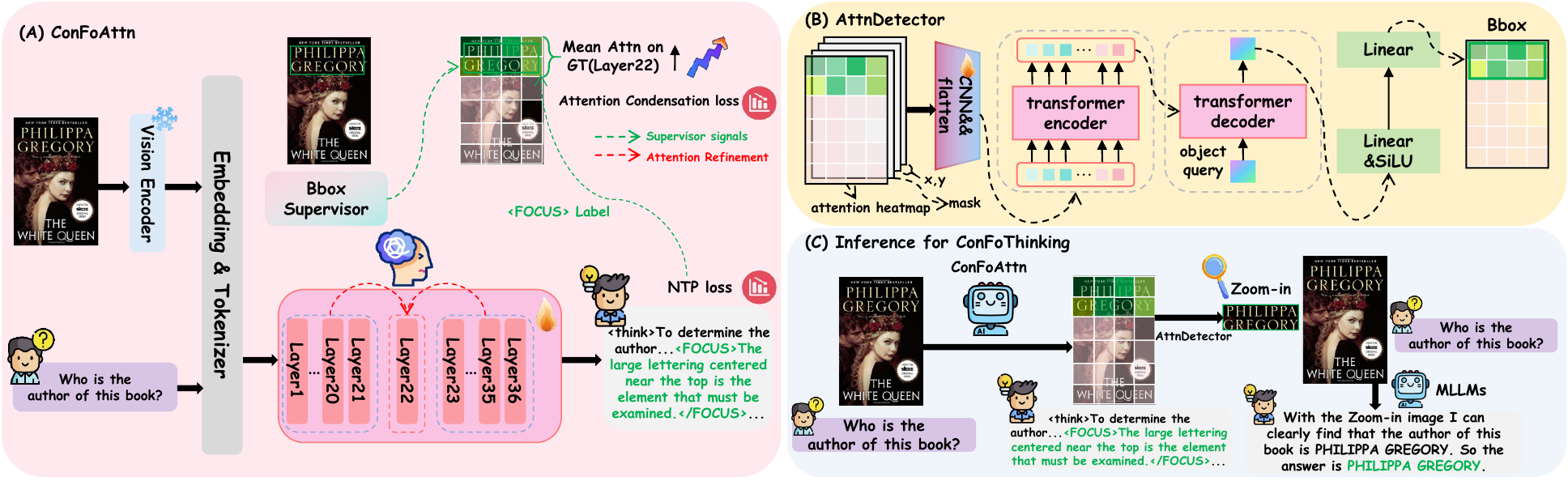}
\caption{\textbf{\Method overview.}
(A) Training ConFoAttn to produce a \texttt{<FOCUS>}...\texttt{</FOCUS>} span and fixed-layer attention heatmaps.
(B) Training AttnDetector to regress ROI boxes from attention heatmaps.
(C) Inference pipeline: generate heatmap, localize and zoom the ROI, then answer with the base MLLM using both the original and zoomed images.  }

  \label{fig:method}
\end{figure*}
\section{Methodology}

\subsection{Prior Coordinate-Output Pipelines}
\label{sec:prior_coordinate_pipeline}

Most existing \emph{thinking-with-images} systems adopt a coordinate-output paradigm to localize key regions.
Given an image $I$ and a question $q$, the model first performs an ROI proposal step by generating bounding-box coordinates (often as a dedicated span of coordinate tokens).
Formally, an MLLM $\mathcal{M}_\theta$ is prompted to produce a predicted ROI box
\begin{equation}
\hat{\mathbf{b}} = f_\theta(I,q), \qquad 
\hat{\mathbf{b}} = (x_1,y_1,x_2,y_2),
\label{eq:prior_roi_proposal}
\end{equation}
where $(x_1,y_1,x_2,y_2)$ are typically normalized to $[0,1]$ (or discretized into a fixed token vocabulary).
Equivalently, the ROI can be viewed as sampled/decoded from a coordinate distribution
$p_\theta(\mathbf{b}\mid I,q)$, and the pipeline uses the decoded $\hat{\mathbf{b}}$ as the region to inspect.

The predicted coordinates are then used to crop and zoom the corresponding image region.
Let $\mathcal{C}(I,\mathbf{b})$ denote a deterministic crop (and optional resize) operator. The pipeline crops the ROI and optionally resizes it to a higher resolution (``zoomed'') so that small objects and fine-grained evidence become more salient:
\begin{equation}
I_{\text{roi}} = \mathrm{Resize}\big(\mathcal{C}(I,\hat{\mathbf{b}})\big).
\label{eq:prior_crop_zoom}
\end{equation}

Finally, the pipeline \textbf{re-queries} an MLLM with the cropped ROI to obtain the final answer
optionally conditioning on both the original image and the crop, i.e., $\hat{a}=g_\theta(I, I_{\text{roi}}, q)$, to retain global context.
This two-stage design can be interpreted as a ``predict coordinates--crop--rethink'' decomposition, where the first pass is responsible for spatial grounding and the second pass focuses on fine-grained reasoning within the localized region.

\subsection{Method Overview}
\label{sec:method_overview}
As shown in Section~\ref{section3.1}, MLLMs often attend to the correct target region in intermediate layers but still output incorrect coordinates, motivating our attention-driven ROI extraction. Section~\ref{section3.3} further reveals that attention extraction is query-sensitive; therefore, we use semantic visual reasoning cues to query attention rather than the raw question. Moreover, Section~\ref{section3.2} shows that where-to-look attention is scattered across layers, so we consolidate it into a designated layer to obtain more stable heatmaps.

Overall, we propose \Method (Fig.~\ref{fig:method}). In the following sections, Section~\ref{section:4.3} describes how we train the model to generate semantically guided visual cues, Section~\ref{section:4.4} explains how we consolidate attention into a fixed layer and derive reliable ROI boxes.

\subsection{Semantically Guided Visual Chain of Thought}\label{section:4.3}

MLLMs are typically trained with next-token prediction (NTP) over discrete symbols, whereas predicting bounding-box coordinates is a continuous output. This mismatch makes accurate coordinate generation difficult to learn. Therefore, in this section, we train \Method to produce semantically guided visual chain of thought instead of directly emitting coordinates.
To equip ConFoAttn with a semantically guided visual chain of thought, we construct a dedicated training dataset for ConFoAttn. Concretely, we construct training pairs in a \emph{focus-centric} format.
Each sample consists of an image, a question, and a rewritten perception CoT that contains a concise semantic cue enclosed by \texttt{<FOCUS>}...\texttt{</FOCUS>}, followed by the final answer.
The \texttt{<FOCUS>} span specifies \emph{what visual evidence to attend to} (a short, verb-containing description), while we additionally attach the corresponding ROI box coordinates as supervision labels, indicating \emph{where the evidence lies}.
Thus, each sample provides (i) a focus-augmented reasoning trace for next-token prediction and (ii) a target box for the attention-condensation objective.
Further construction details are provided in Appendix~\ref{app:data_construction}.

During training, we employ the next-token prediction loss $L_{\text{NTP}}$ to equip the model with semantically grounded visual reasoning. This loss teaches the model to generate object-centric semantic cues from the question, forming the \texttt{<FOCUS>} visual hints. Taking Fig.~\ref{fig:method}(a) as an example, given the image and question, the model generates a semantically guided visual chain of thought, as illustrated in the bottom-right of Fig.~\ref{fig:method}(a): \texttt{<FOCUS>The large lettering centered near the top is the element that must be examined.</FOCUS>} \ldots. This \texttt{<FOCUS>} cue is then used as the query for subsequent attention extraction.


\subsection{Designated-Layer Attention Aggregation}
\label{section:4.4}
Since the model’s internal attention is dispersed across layers, we consolidate it into a designated layer to obtain a more stable where-to-look signal.
We select the designated layer $\ell$ by evaluating the \emph{base} (unfine-tuned) backbone on the validation set, VisCoT (not our test dataset in comparison): we compute $s^{(\ell)}(\mathcal{R}_{\text{tar}})$ (Eq.~\ref{eq:attn_mean_ratio}) at each layer and choose the layer with the highest average score across samples. For Qwen3-VL-8B (36 layers), this procedure selects Layer~22; we apply the same criterion to other backbones.




For attention condensation, we extract the text-to-image attention at the fixed layer $\ell$, using tokens inside \texttt{<FOCUS>}...\texttt{</FOCUS>} as queries and image tokens as keys. We aggregate attention across query tokens and heads (Eq.~\ref{eq:attn_map_aggregation}) to obtain a heatmap over image tokens. As illustrated in Fig.~\ref{fig:method}(a), given a generated focus span (e.g., \texttt{<FOCUS>The large lettering centered near the top is the element that must be examined.</FOCUS>}), we encourage the heatmap to concentrate inside the dataset-provided target box by minimizing $\mathcal{L}_{\mathrm{AC}}$ in Eq.~\ref{eq:Lac}.

\begin{equation}\label{eq:Lac}
\mathcal{L}_{\mathrm{AC}} \;=\; -\log \big( s^{(\ell)}(\mathcal{R}_{\text{tar}}) \big).
\end{equation}

Finally, we combine $\mathcal{L}_{\mathrm{NTP}}$ and $\mathcal{L}_{\mathrm{AC}}$ with a weighting factor to form the overall objective $\mathcal{L}_{\text{total}}$, which is used to optimize the model in Eq.~\ref{eq:Ltotal}.

\begin{equation}\label{eq:Ltotal}
\mathcal{L}_{\text{total}}=\mathcal{L}_{\mathrm{NTP}}+\alpha\mathcal{L}_{\mathrm{AC}} \, .
\end{equation}

The trained ConFoAttn offers two advantages. First, it generates a compact perception CoT and explicitly marks a focus span, so heatmaps are queried from a small, semantically relevant token set. Second, where-to-look attention becomes more concentrated in the designated layer instead of being scattered across layers, yielding more stable attention heatmaps for ROI mining.

After obtaining the attention heatmap, a key question is how to convert it into a final bounding box. We address this by training an object detector, \textbf{AttnDetector}, which takes the heatmap as input and outputs ROI coordinates.
To train AttnDetector, we construct paired heatmap--box supervision. Specifically, we run the trained ConFoAttn on the ConFoAttn dataset, extract the designated-layer attention heatmap queried by the \texttt{<FOCUS>}...\texttt{</FOCUS>} span, and use the dataset-provided target box as supervision, forming the AttnDetector training set. We then train AttnDetector as a transformer-based detector with an $\mathcal{L}_1$ loss and a GIoU loss; details are provided in Appendix~\ref{app:AttnDetector}. 

\paragraph{Inference for ConFoThinking}
During inference, we follow the standard thinking-with-images pipeline: predict coordinates, crop, and rethink. Specifically, ConFoAttn and AttnDetector jointly predict the ROI box; we then crop and zoom the region, and feed the zoomed crop together with the original image into the backbone model to produce the final answer.

\begin{table*}[h]
\centering
\caption{Results on high-resolution, OCR, and general VQA benchmarks. For \textbf{Ours}, the \textcolor{darkgreen}{$\uparrow$ value} denotes the absolute gain over the corresponding base backbone (e.g., \Method(Qwen2.5-VL-7B) vs.\ Qwen2.5-VL-7B).}
\label{Table:Comparison_experiments}
{\fontsize{8}{9}\selectfont 
\setlength{\tabcolsep}{1.5pt} 
\setlength{\arrayrulewidth}{0.5pt} 
\renewcommand{\arraystretch}{1.08}

\begin{tabular}{l|ccc|ccc|ccc|c|c}
\toprule
\multirow{3}{*}{Model} &
\multicolumn{9}{c|}{High-resolution} &
\multicolumn{1}{c|}{OCR} &
\multicolumn{1}{c}{General} \\
\cline{2-12}

&
\multicolumn{3}{c|}{V* Bench} &
\multicolumn{3}{c|}{HRBench-4K} &
\multicolumn{3}{c|}{HRBench-8K} &
\multirow{2}{*}{$\text{InfoVQA}_{\text{Val}}$} &
\multirow{2}{*}{GQA} \\
\cline{2-10}

&
Attr. & Spatial & Overall &
FSP & FCP & Overall &
FSP & FCP & Overall &
& \\
\midrule

\multicolumn{9}{l}{\textit{Open-source Models}} \\
\midrule
Qwen2.5-VL-7B & 78.2 & 73.6 & 76.4 & 85.3 & 52.3 & 68.8 & 78.8 & 51.8 & 65.3 & 77.9 & 62.1 \\
Qwen3-VL-4B & 80.0 & 75.0 & 78.0 & 86.0 & 62.5 & 74.3 & 82.3 & 58.0 & 70.1 & 79.3 & 69.3 \\
Qwen3-VL-8B & 86.1 & 81.6 & 84.3 & 88.3 & 64.5 & 76.4 & 86.3 & 58.0 & 72.1 & 81.7 & 71.3 \\
InternVL3.5-8B & 82.6 & 73.7 & 79.1 & 85.3 & 60.8 & 73.2 & 84.3 & 57.3 & 70.8 & 79.2 & 68.2 \\
LLaVA-OneVision-1.5-8B & 79.1 & 76.3 & 78.0 & 85.8 & 58.2 & 72.0 & 82.0 & 56.5 & 69.3 & 78.1 & 67.7  
\\
\midrule

\multicolumn{9}{l}{\textit{Thinking with Images}} \\
\midrule
Qwen3-VL-4B (with tools) &85.2&80.3&83.2&84.3&62.3&73.3&81.5&61.0&71.3&81.1 & 70.2 \\
Qwen3-VL-8B (with tools) &88.7&85.5&86.4&85.0&65.5&75.3&83.8&66.0&74.9&83.1&71.9\\
Pixel-Reasoner & 85.2 & 82.9 & 84.3 & 86.3 & 61.8 & 74.0 & 80.1 & 59.8 & 66.9 & 79.8 & 63.4 \\
Visual Sketchpad & 81.7 & 77.6 & 80.1 & 81.3 & 57.8 & 69.5 & 75.3 & 53.0 & 64.1 & 75.1 & 63.2 \\
SEAL & 78.3 & 68.7 & 74.9 & 71.5 & 49.0 & 60.3 & 70.1 & 47.5 & 58.8 & 70.2 & 59.8 \\
ZoomEye & 93.9 & 85.5 & 90.6 & 84.2 & 55.0 & 69.6 & 88.5 & 50.0 & 69.3 & 76.2 & 62.1 \\
ICoT & 63.5 & 56.6 & 60.7 & 65.5 & 55.0 & 60.3 & 58.0 & 48.5 & 53.3 & 58.1 & 50.7 \\
ViCrop &65.2&57.9&62.3& --&--&--&--&--&--&--&--\\
DyFo & 80.0 & 82.9 &81.2 & --&--&--&--&--&--&--&--
\\
\midrule
\multicolumn{9}{l}{\textit{Ours}} \\
\midrule
\textbf{\Method(Qwen2.5-VL-7B)} & $\textbf{90.4}_{\textcolor{darkgreen}{\uparrow 12.2}}$ & $\textbf{81.5}_{\textcolor{darkgreen}{\uparrow 7.9}}$ & $\textbf{86.9}_{\textcolor{darkgreen}{\uparrow 10.5}}$ & $\textbf{91.8}_{\textcolor{darkgreen}{\uparrow 6.5}}$ & $\textbf{56.8}_{\textcolor{darkgreen}{\uparrow 4.5}}$ & $\textbf{74.3}_{\textcolor{darkgreen}{\uparrow 5.5}}$ & $\textbf{85.0}_{\textcolor{darkgreen}{\uparrow 6.2}}$ & $\textbf{57.0}_{\textcolor{darkgreen}{\uparrow 5.2}}$ & $\textbf{71.0}_{\textcolor{darkgreen}{\uparrow 5.7}}$ & $\textbf{82.1}_{\textcolor{darkgreen}{\uparrow 4.2}}$ & $\textbf{66.7}_{\textcolor{darkgreen}{\uparrow 4.6}}$ \\
\textbf{\Method(Qwen3-VL-4B)} & $\textbf{93.9}_{\textcolor{darkgreen}{\uparrow 13.9}}$ & $\textbf{86.8}_{\textcolor{darkgreen}{\uparrow 11.8}}$ & $\textbf{91.1}_{\textcolor{darkgreen}{\uparrow 13.1}}$ & $\textbf{87.8}_{\textcolor{darkgreen}{\uparrow 1.8}}$ & $\textbf{68.0}_{\textcolor{darkgreen}{\uparrow 5.5}}$ & $\textbf{77.9}_{\textcolor{darkgreen}{\uparrow 3.6}}$ & $\textbf{83.5}_{\textcolor{darkgreen}{\uparrow 1.2}}$ & $\textbf{66.8}_{\textcolor{darkgreen}{\uparrow 8.8}}$ & $\textbf{75.1}_{\textcolor{darkgreen}{\uparrow 5.0}}$ & $\textbf{86.4}_{\textcolor{darkgreen}{\uparrow 7.1}}$ & $\textbf{74.2}_{\textcolor{darkgreen}{\uparrow 4.9}}$\\
\textbf{\Method(Qwen3-VL-8B)} &$\textbf{94.8}_{\textcolor{darkgreen}{\uparrow 8.7}}$ & $\textbf{88.1}_{\textcolor{darkgreen}{\uparrow 6.5}}$ & $\textbf{92.1}_{\textcolor{darkgreen}{\uparrow 7.8}}$ & $\textbf{92.8}_{\textcolor{darkgreen}{\uparrow 4.5}}$ & $\textbf{67.3}_{\textcolor{darkgreen}{\uparrow 2.8}}$ & $\textbf{80.0}_{\textcolor{darkgreen}{\uparrow 3.6}}$ & $\textbf{87.3}_{\textcolor{darkgreen}{\uparrow 1.0}}$ & $\textbf{68.3}_{\textcolor{darkgreen}{\uparrow 10.3}}$ & $\textbf{77.8}_{\textcolor{darkgreen}{\uparrow 5.7}}$ & $\textbf{87.9}_{\textcolor{darkgreen}{\uparrow 6.2}}$ & $\textbf{74.9}_{\textcolor{darkgreen}{\uparrow 3.6}}$\\
\bottomrule
\end{tabular}
} 
\end{table*}

\section{Experiments}


This section evaluates \Method from three perspectives. We first compare \Method against strong open-source MLLMs and representative thinking-with-images pipelines on diverse benchmarks to verify its overall effectiveness. We then conduct ablations to quantify the contribution of each component. Finally, we provide interpretability analyses to show that \Method indeed consolidates where-to-look attention into a fixed layer.

\subsection{Implementation Details}
\label{sec:impl}

All training and inference are conducted on 8$\times$NVIDIA A800 (80GB) GPUs. We fine-tune ConFoAttn with $\alpha{=}0.003$, learning rate $=1\times10^{-5}$, batch size $=2$, and 2 epochs; the vision encoder is frozen while the connector and LLM are updated.
\subsection{Comparison Experiments}

We evaluate \Method on five multimodal benchmarks covering high-resolution perception, OCR understanding, and general VQA: V*~\cite{wu2024v}, HR-Bench 4K/8K~\cite{wang2025divide}, InfoVQA~\cite{mathew2022infographicvqa}, and GQA~\cite{hudson2019gqa}. V*, HR-Bench 4K, and HR-Bench 8K stress-test fine-grained perception under large images; InfoVQA focuses on OCR; and GQA evaluates general visual reasoning. We compare against (i) open-source MLLMs, including Qwen2.5-VL~\cite{Qwen2.5-VL}, Qwen3-VL\mbox
{~\cite{bai2511qwen3}}, InternVL3.5~\cite{wang2025internvl3_5}, and LLaVA-OneVision-1.5~\cite{an2025llava}, and (ii) thinking-with-images pipelines, including Qwen3-VL with tools~\cite{bai2511qwen3}, Pixel-Reasoner~\cite{wang2025pixel}, Visual Sketchpad~\cite{hu2024visual}, SEAL~\cite{wu2024v}, ZoomEye~\cite{shen2025zoomeye}, ICoT~\cite{gao2025interleaved}, ViCrop~\cite{zhang2025mllms}, and DyFo~\cite{li2025dyfo}. We additionally report inference latency on V*: \Method (Qwen3-VL-8B) takes 12.1\,s per sample on an A800 GPU, while ZoomEye takes 49.8\,s due to its multi-step, search-style region exploration. This indicates that although ZoomEye performs well on V*, its inference cost is prohibitively high, making it impractical. In contrast, \Method achieves SOTA performance with a reasonable inference budget, running about $\sim$5$\times$ faster than ZoomEye.
Notably, we use the validation set, VisCoT, only to select the layer for attention aggregation; VisCoT is not included in the evaluation benchmarks for the comparison experiments, and thus there is no risk of data leakage.
Results are summarized in Table~\ref{Table:Comparison_experiments}.

\subsection{Ablation Experiments}


\paragraph{Ablation I: Effect of NTP and Attention Condensation.}
We ablate the training objectives for ConFoAttn on V* with two backbones (Qwen2.5-VL-7B and Qwen3-VL-8B). Table~\ref{tab:ablation_loss} compares: (i) \emph{w/o NTP \& AC}, which performs direct VQA on the full image without crop+zoom; (ii) \emph{NTP only}, which enables our crop+zoom pipeline and trains ConFoAttn with $\mathcal{L}_{\mathrm{NTP}}$; and (iii) \emph{NTP + AC}, which adds the attention-condensation loss $\mathcal{L}_{\mathrm{AC}}$. Enabling crop+zoom with NTP-only improves accuracy from 76.4/84.3 to 81.2/88.0, and adding $\mathcal{L}_{\mathrm{AC}}$ further reaches 86.9/92.1. This shows that NTP helps produce informative \texttt{<FOCUS>} cues for ROI mining, while attention condensation stabilizes where-to-look signals and yields more reliable crops.

\begin{table}[h]
\centering
\caption{Ablation I: effect of attention-condensation training.}
\label{tab:ablation_loss}
\footnotesize
\setlength{\tabcolsep}{5pt}
\renewcommand{\arraystretch}{1.1}
\begin{tabular}{lcc}
\toprule
Method & Qwen2.5-VL-7B & Qwen3-VL-8B \\
\midrule
w/o NTP \& AC & 76.4 & 84.3 \\
NTP only                 & 81.2 & 88.0 \\
NTP + AC & \textbf{86.9} & \textbf{92.1} \\
\bottomrule
\end{tabular}
\end{table}

\paragraph{Ablation II: Which Text to Query Attention.}
We evaluate how the choice of \emph{text query} affects attention-based ROI mining on \textsc{V*}.
We keep the trained models and the overall inference pipeline fixed, and only vary which text span is used to query text-to-image attention when forming the heatmap: (i) the question, (ii) all generated tokens, or (iii) the tokens inside \texttt{<FOCUS>}...\texttt{</FOCUS>}.
As shown in Table~\ref{tab:ablation_text}, querying with all generated text performs worst, suggesting that redundant reasoning context introduces semantic noise into attention extraction.
Question-only queries perform better but can still be underspecified.
Using the \texttt{<FOCUS>} span yields the best accuracy on both backbones (e.g., $92.1$ vs.\ $89.0$ on Qwen3-VL-8B), supporting our design choice of extracting attention from concise, semantically refined cues.

\begin{table}[h]
\centering
\caption{Ablation II: text query for attention extraction.}
\label{tab:ablation_text}
\footnotesize
\setlength{\tabcolsep}{5pt}
\renewcommand{\arraystretch}{1.1}
\begin{tabular}{lcc}
\toprule
Text Query  & Qwen2.5-VL-7B & Qwen3-VL-8B \\
\midrule
Question-only         & 85.9 & 89.0 \\
All generated text  & 80.1  & 84.3 \\
\texttt{<FOCUS>} span   & \textbf{86.9} & \textbf{92.1} \\
\bottomrule
\end{tabular}
\end{table}

\paragraph{Ablation III: Single-layer vs.\ Neighborhood Layer Aggregation (Qwen3-VL-8B).}
Although \Method consolidates where-to-look attention into a designated layer, a natural alternative is to average a small neighborhood of layers to mitigate residual layer-wise variation.
We evaluate this idea on Qwen3-VL-8B (36 layers) by comparing three extraction sets: a single layer $\mathcal{S}_{1}=\{22\}$, a 3-layer window $\mathcal{S}_{3}=\{21,22,23\}$, and a 5-layer window $\mathcal{S}_{5}=\{20,21,22,23,24\}$.
For a fair comparison, each variant uses the \emph{same} layer set $\mathcal{S}$ in both training (to define the attention-condensation supervision) and inference (to form the heatmap input to AttnDetector); the only difference is whether the heatmap is extracted from one layer or averaged across a local window.

As reported in Table~\ref{tab:ab_layer_window_qwen3vl8b}, single-layer extraction achieves the best accuracy (92.1), while neighborhood averaging over $\mathcal{S}_{3}$ and $\mathcal{S}_{5}$ degrades performance (91.1 and 90.6).
This suggests that once where-to-look attention is explicitly condensed into a designated layer, averaging across neighboring layers can reintroduce off-target signals and blur the ROI heatmap, leading to less precise crops and lower downstream accuracy.
\vspace{-10pt}

\begin{table}[t]
\centering
\caption{Ablation III on \textsc{V*} (Qwen3-VL-8B): single-layer extraction vs. neighborhood layer aggregation (the same layer set is used for training and inference).}
\label{tab:ab_layer_window_qwen3vl8b}
{\fontsize{8}{9}\selectfont
\setlength{\tabcolsep}{4pt}
\renewcommand{\arraystretch}{1.05}
\begin{tabular}{lcc}
\toprule
\textbf{Extraction layers} & \textbf{Train} & \textbf{V*} \\
\midrule
Single layer & $\{22\}$ & 92.1 \\
3-layer window & $\{21,22,23\}$ & 91.1 \\
5-layer window & $\{20,21,22,23,24\}$ & 90.6 \\
\bottomrule
\end{tabular}}
\end{table}
\subsection{Attention Condensation Analysis}
\begin{figure}[!h]
  \centering
  \includegraphics[width=0.48\textwidth]{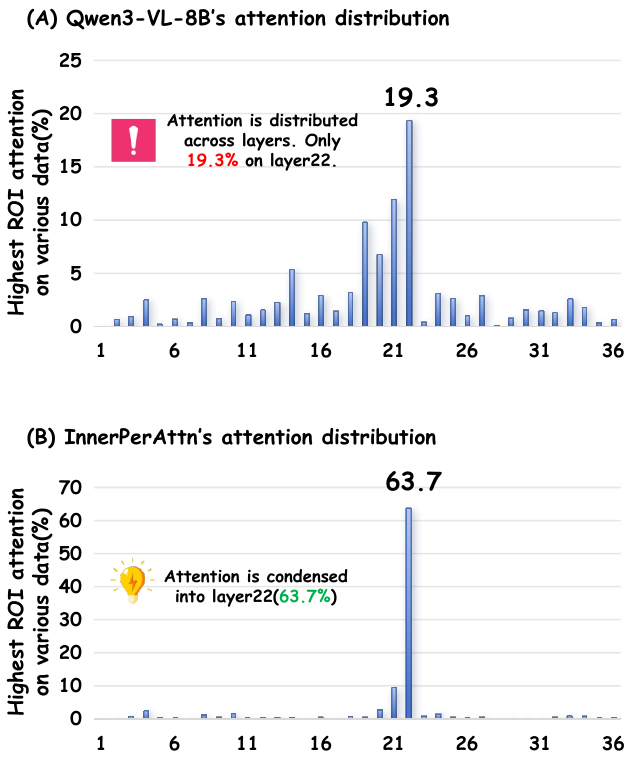}
  \caption{Comparison for attention condensation.}
  \label{fig:Atten_condensation}
  \vspace{-10pt}
\end{figure}

Beyond accuracy gains, we verify that attention-condensation training makes \emph{fixed-layer} attention extraction reliable. On the validation set, VisCoT, we compute the text-to-image attention induced by the generated \texttt{<FOCUS>}...\texttt{</FOCUS>} span at \emph{every} decoder layer and identify the peak-attention layer for each sample as $\ell^{\star}=\arg\max_{\ell} s^{(\ell)}(\mathcal{R}_{\text{tar}})$ (Eq.~(3)). 
Fig.~\ref{fig:Atten_condensation} shows that after attention-condensation training, 63.7\% of samples peak at the designated Layer~22, whereas the base Qwen3-VL-8B peaks at Layer~22 for only 19.3\% of samples. 
This shift indicates that where-to-look signals become substantially more concentrated in the chosen layer, which validates fixed-layer attention extraction and supports stable ROI mining in \Method.
\section{Conclusion}
This paper addresses a key bottleneck in Thinking with Images: reliable ROI localization for evidence-grounded reasoning. Our empirical study identifies three failure modes: (i) coordinate-output pipelines suffer from a grounding--perception mismatch, (ii) attention-driven cropping is unstable due to fragmented where-to-look signals across layers, and (iii) attention-based ROI mining is query-sensitive, where using raw questions as queries can yield diffuse and less ROI-specific attention than semantically guided cues. To address these issues, we propose \Method, which uses concise \texttt{<FOCUS>} cues to query text-to-image attention with reduced semantic noise, condenses where-to-look attention into a designated intermediate layer, and employs AttnDetector to convert the resulting heatmaps into ROI boxes for zoom-and-answer with the base MLLM. Experiments on five benchmarks (V*, HR-Bench 4K/8K, InfoVQA, and GQA) show consistent gains, demonstrating a simple and effective way to obtain robust ROIs without relying on brittle coordinate generation.

\paragraph{Limitations.}
We train attention condensation and AttnDetector with box supervision; exploring weaker (e.g., pseudo-label or self-training) supervision may further broaden applicability.
Additionally, we focus on VQA-style benchmarks, and future work can evaluate \Method on other multimodal tasks where localized evidence acquisition is also important.
\clearpage

\nocite{langley00}

\bibliography{example_paper}
\bibliographystyle{icml2026}

\newpage
\appendix
\onecolumn
\section*{Appendix}
\section{Grounding-Perception Mismatch for Pixel-Reasoner}\label{Appendix:Grounding-Perception_Mismatch}

We analyze whether Pixel-Reasoner’s explicit ROI prediction faithfully reflects its internal \emph{where-to-look} signals, following the same protocol as Section~\ref{section3.1}.
For each sample, let $R_{\text{pred}}$ denote the set of tokens in ROI box predicted by Pixel-Reasoner and $R_{\text{GT}}$ those in the ground-truth target box.
We define a \emph{grounding-error} case when $\mathrm{IoU}(R_{\text{pred}}, R_{\text{GT}}) < \tau$ (we use $\tau=0.1$), and focus on the subset where the final answer is incorrect.
We then measure layer-wise attention concentration on a region $R$ using the normalized mean-attention ratio $s^{(\ell)}(R)$ (Eq.~\ref{eq:attn_mean_ratio} in the main paper), and compare $s^{(\ell)}(R_{\text{GT}})$ against $s^{(\ell)}(R_{\text{pred}})$ across layers.

\begin{figure}[h]
    \centering
    \includegraphics[width=0.6\linewidth]{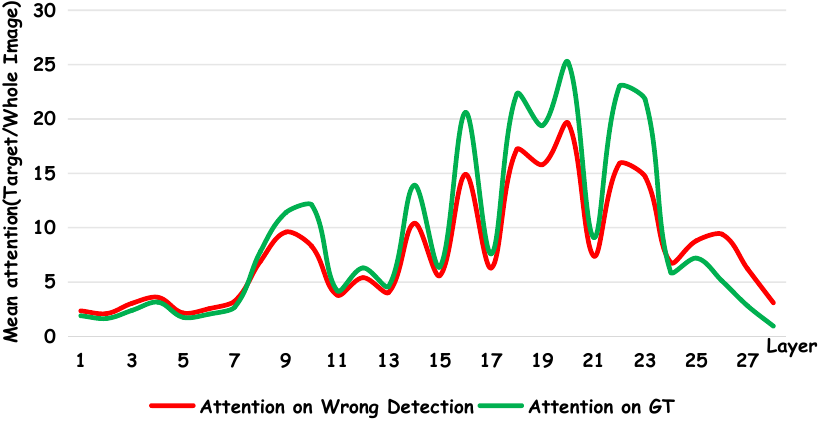}
    \caption{Layer-wise comparison of attention concentration on $R_{\text{GT}}$ vs.\ $R_{\text{pred}}$ for Pixel-Reasoner over grounding-error cases ($\mathrm{IoU}<0.1$).}
    \label{fig:app_pixelreasoner_gpm}
\end{figure}

We observe the same stage-wise pattern as in coordinate-output tool pipelines (Fig.~\ref{fig:app_pixelreasoner_gpm}).
In intermediate vision--language fusion layers, attention concentrates more on the true target region, i.e., $s^{(\ell)}(R_{\text{GT}}) > s^{(\ell)}(R_{\text{pred}})$, indicating that Pixel-Reasoner has already localized the correct evidence internally.
However, in later decoding layers that are closer to ROI/coordinate emission, attention drifts toward the predicted (but incorrect) region, and $s^{(\ell)}(R_{\text{pred}})$ can surpass $s^{(\ell)}(R_{\text{GT}})$.
This demonstrates a \emph{grounding--perception mismatch} in Pixel-Reasoner: even when the model ``knows where to look'' during fusion, the decoded ROI can deviate due to late-layer attention drift, leading to inaccurate cropping and ultimately wrong answers.
\paragraph{Implication.}
These results further support our motivation that explicitly generating ROI coordinates is fragile: the model’s internal perception signal can be correct, yet the decoded ROI can be unreliable.
This motivates extracting ROIs from stabilized internal attention rather than directly relying on coordinate decoding.

\section{Peak-Attention Layer Dispersion on Other Backbones}
\label{app:peak-layer-other-backbones}

\paragraph{Setup.}
Following Section~\ref{section3.2}, we define the peak-attention layer for each sample as
$\ell^\star = \arg\max_{\ell\in\{1,\dots,L\}} s^{(\ell)}(R_{\text{tar}})$,
and report the empirical distribution of $\ell^\star$ on the validation set, VisCoT, for additional backbones.

\begin{figure}[t]
    \centering
    \includegraphics[width=\linewidth]{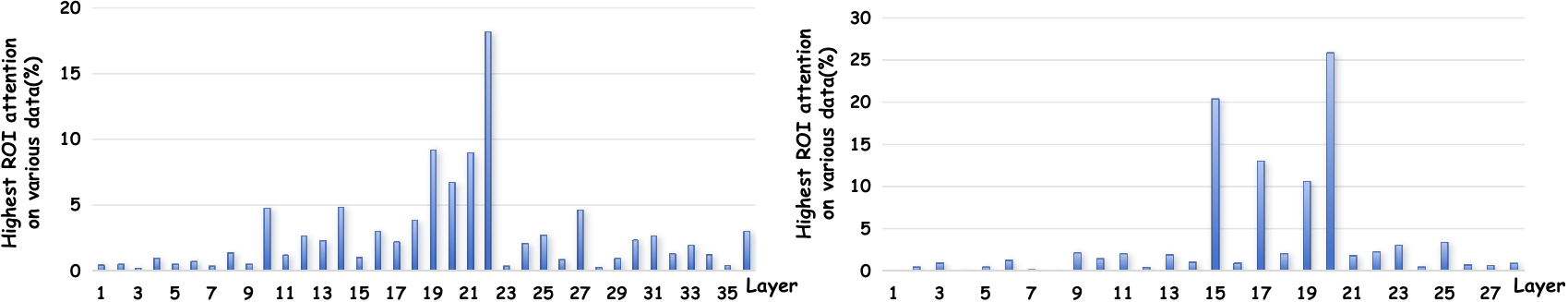}
    \caption{Distribution of the peak-attention layer $\ell^\star$ on the Validation set, VisCoT.
    Left: Qwen3-VL-4B (mode at Layer~22). Right: Qwen2.5-VL-7B (mode at Layer~20).
    Despite the existence of a modal layer, the distribution remains broadly dispersed across layers.}
    \label{fig:app_attn_distribution}
\end{figure}

\paragraph{Results.}
Figure~\ref{fig:app_attn_distribution} shows that the peak-attention layer $\ell^\star$ is highly sample-dependent and remains broadly dispersed across layers for both backbones.
Although a single \emph{modal} layer exists, it is far from dominant: for Qwen3-VL-4B, Layer~22 is the most frequent peak, yet it accounts for only $18.2\%$ of the samples, with the remaining $81.8\%$ peaking at other layers; for Qwen2.5-VL-7B, the mode shifts to Layer~20, but it still covers only $25.9\%$ of the samples, leaving $74.1\%$ distributed elsewhere.
These results indicate that the ``best'' layer for attention-based ROI mining is not stable across instances, even when a highest-frequency layer can be identified, which further motivates consolidating where-to-look attention into a designated layer.

\section{Details for Dataset Construction}
\label{app:data_construction}

\paragraph{Overview.}
To train \textsc{ConFoAttn}, we construct an \textsc{ConFoAttn} dataset by rewriting \emph{coordinate-centric} reasoning traces into \emph{focus-centric} ones.
The raw sources include (i) VGR and (ii) the Pixel-Reasoner SFT dataset.
In total, the dataset contains 48,640 VQA pairs, including 47,575 samples from VGR and 1,065 samples from Pixel-Reasoner SFT.
We use GPT-5 as the rewriting teacher and keep the original box coordinates as supervision labels.%

\subsection{VGR: Coordinate-to-Focus Distillation}
\label{app:vgr_distill}

\paragraph{Input format.}
Each VGR example contains an image $I$, a question $q$, and an original assistant response that may include $N$ region markers of the form
\texttt{<SOT>[x1,y1,x2,y2]<EOT><image>} (possibly with multiple regions).

\paragraph{Teacher rewriting objective.}
Given the original response as \emph{guidance}, GPT-5 rewrites the assistant output into the following strict structure:
\begin{verbatim}
<think>
...rewritten reasoning with one or more <FOCUS> blocks inserted in-line...
</think>
<answer>
...final answer...
</answer>
\end{verbatim}
The rewriting follows three key constraints:

\begin{itemize}
  \item \textbf{FOCUS cardinality.}
  If the original response contains $N$ \texttt{<SOT>} region markers, the rewritten reasoning must contain exactly $N$ \texttt{<FOCUS>} blocks.
  If the original response contains no region markers, we enforce exactly one \texttt{<FOCUS>} block.
  \item \textbf{FOCUS placement.}
  The $i$-th \texttt{<FOCUS>} block must be inserted immediately after the point where the rewritten reasoning \emph{first} refers to the corresponding region/object, and the order must match the \texttt{<SOT>} order.
  \item \textbf{FOCUS content.}
  Each \texttt{<FOCUS>} block contains exactly one sentence (must include a verb) that \emph{only} identifies the region/object to attend to (semantic/relative description).
  It must not include coordinates, OCR text, colors, counts, or any words that directly reveal the final answer.
\end{itemize}

\paragraph{Labels.}
We retain the original bounding box coordinates associated with each \texttt{<SOT>} region as target-box labels.
Thus, each \texttt{<FOCUS>} span is paired with its target box for (i) attention-condensation supervision during \textsc{ConFoAttn} training and (ii) heatmap-to-box supervision when constructing the \textsc{AttnDetector} dataset.

\subsection{Pixel-Reasoner SFT: Two Distillation Variants}
\label{app:pixelreasoner_distill}

Pixel-Reasoner SFT traces often contain explicit region proposals and may include iterative trial-and-error.
To better model these two regimes, we split the Pixel-Reasoner subset into Single-Pass and Recrop groups and apply different distillation rules.

\subsubsection{Single-Pass Distillation}
\label{app:pixelreasoner_singlepass}

\paragraph{When used.}
We categorize an example as Single-Pass when the original guidance indicates a direct (non-iterative) successful localization path (typically a single round of region grounding without repeated failures).

\paragraph{Rewriting constraints.}
We enforce the same \texttt{<think>/<answer>} structure and the same FOCUS cardinality/placement/content rules as in VGR:
if the guidance contains $N$ region markers, we produce exactly $N$ \texttt{<FOCUS>} blocks, aligned in the same order.

\paragraph{Style constraints (no ``zoom'').}
To avoid tool-centric phrasing, the teacher is forbidden to use image-manipulation terms such as
\emph{``zoom in'', ``crop'', ``cut'', ``resize'', ``scale'', ``tool'', ``selection''}.
Instead, it must use natural visual-attention language (e.g., ``I examine...'', ``I focus my attention on...'', ``Looking closer at...'').

\paragraph{Answer normalization.}
If the original answer uses a boxed format (e.g., \texttt{\textbackslash boxed\{A\}}), the teacher must remove the wrapper and output only the literal answer text inside \texttt{<answer>} (no coordinates or boxes).

\paragraph{Labels.}
We keep the target box(es) linked to each region marker as supervision labels, identical to VGR.

\subsubsection{Recrop Distillation}
\label{app:pixelreasoner_recrop}

\paragraph{When used.}
We categorize an example as Recrop when the original guidance exhibits a \emph{trial-and-error} process:
multiple region attempts are made before the final successful one.

\paragraph{Key idea: rewrite as ``right on the first try''.}
The teacher must ignore failed attempts and only keep the final successful region that provides the evidence for the answer.
Concretely, it synthesizes a direct reasoning path that:
(i) briefly scans global context,
(ii) identifies the target region,
(iii) observes evidence within that region,
(iv) concludes the answer.

\paragraph{Strict reasoning flow.}
Inside \texttt{<think>}, we enforce the following order:
\begin{enumerate}
  \item \textbf{Context:} briefly describe global scan of the image.
  \item \textbf{Focus (Action):} identify the \emph{final successful} target region and insert exactly one \texttt{<FOCUS>} block here.
  \item \textbf{Observation (Result):} after \texttt{<FOCUS>}, describe what is observed in that region (OCR/text/numbers/details allowed \emph{outside} \texttt{<FOCUS>}).
  \item \textbf{Conclusion:} derive the final answer.
\end{enumerate}

\paragraph{Style \& answer constraints.}
We apply the same no-``zoom'' style constraints and no boxed answer normalization as Single-Pass.

\paragraph{Labels.}
Since Recrop distillation collapses the history into a single successful localization, we keep only the bounding box of the \emph{final successful} region as the target-box label for this sample.

\subsection{Prompts Used for Rewriting}
\label{app:data_construction_prompt}

We provide the exact GPT-5 prompts for:
(i) VGR distillation, (ii) Pixel-Reasoner Single-Pass distillation, and (iii) Pixel-Reasoner Recrop distillation
in the supplementary PDF.

\begin{tcolorbox}[
  breakable,
  title={Prompt For Constructing Dataset From VGR},
  colback=gray!10,
  colframe=gray!60,
  fonttitle=\bfseries,
]
You must rewrite the original assistant reasoning and output in exactly the following structure and nothing else: \\
\\
\texttt{<think>} \\
\texttt{...your rewritten reasoning with one or more <FOCUS> blocks inserted in-line...} \\
\texttt{</think>} \\
\\
\texttt{<answer>} \\
\texttt{...final answer...} \\
\texttt{</answer>} \\
\\

Your task: \\
\\

1. Use the original assistant response (with \texttt{<SOT>...[x1,y1,x2,y2]...<EOT><image>} markers) as guidance. \\
2. Rewrite the reasoning in your own words, but keep the same general logic and final answer. \\
3. Insert \texttt{<FOCUS>} blocks at the correct places inside the reasoning according to the rules below. \\
\\

FOCUS placement rules: \\
\\
- If the original assistant response contains $N$ region markers of the form \texttt{<SOT>[...coordinates...]<EOT><image>}, you MUST produce exactly $N$ \texttt{<FOCUS>...</FOCUS>} blocks. \\
- The order of the \texttt{<FOCUS>} blocks MUST follow the order of the \texttt{<SOT>} regions in the original response. \\
- For each region, insert its \texttt{<FOCUS>} block immediately after the moment in your rewritten reasoning where you FIRST start talking about that specific object/region. \\
- \texttt{<FOCUS>} blocks are NOT required to appear at the beginning of \texttt{<think>}. They must appear where the region first becomes relevant in the reasoning. \\
- If the original assistant response has NO \texttt{<SOT>} region markers, you must produce exactly one \texttt{<FOCUS>} block, inserted at the point where you first look at the key evidence in the image/diagram/text.\\
\\

Content rules for each \texttt{<FOCUS>} block: \\
\\
- Each \texttt{<FOCUS>} block must contain exactly ONE sentence. \\
- The sentence must be a complete sentence and must contain a verb. \\
- The sentence must ONLY identify which object or region you are focusing on for that part of the reasoning. \\
- Use relative or semantic descriptions that distinguish the region, e.g.:\\
\hspace*{1.5em} - ``The stuffed animal on the left is the first object that must be examined."\\
\hspace*{1.5em} - ``The table in front of the seating area is the object that must be examined."\\
\hspace*{1.5em} - ``The bar for the female category is the second item that must be examined."\\
\hspace*{1.5em} - ``The person wearing a jacket near the chairs is the element that must be examined."\\

- The sentence must NOT include any of the following:\\
\hspace*{1.5em} - colors, textures, materials, patterns (e.g., ``red", ``wooden", ``pixelated")\\
\hspace*{1.5em} - exact numbers, percentages, or counts (e.g., ``44\%", ``two", ``three")\\
\hspace*{1.5em} - OCR or label text content (book titles, axis labels, category names, etc.)\\
\hspace*{1.5em} - words that directly appear in the final answer\\
\hspace*{1.5em} - direct descriptive attributes that already reveal why the answer is correct\\
- The sentence must NOT contain meta-instructional content, such as:\\
\hspace*{1.5em} - ``You must look at..."\\
\hspace*{1.5em} - ``This region must be examined to answer the question as instructed..."\\
- Different \texttt{<FOCUS>} blocks must not be identical; each must uniquely point to its own region/object.\\
- Do NOT mention \texttt{<SOT>} or \texttt{<EOT>} or coordinates anywhere.\\
\\

Reasoning rules (content outside \texttt{<FOCUS>}):\\
\\

- The rest of the text inside \texttt{<think>} (outside \texttt{<FOCUS>} blocks) is your normal reasoning.\\
- Use the objects/regions identified by the \texttt{<FOCUS>} blocks as evidence.\\
- Describe observable details (colors, shapes, text, numbers, positions, etc.) OUTSIDE of \texttt{<FOCUS>}, and explain how they lead to the final answer.\\
- You MAY describe colors, text, numbers, etc. in the reasoning, but NEVER inside a \texttt{<FOCUS>} block.\\
- Do NOT copy the original reasoning verbatim; paraphrase it, but keep the same logical meaning.\\
- Do NOT add any meta discussion about prompts, rules, or instructions.\\
\\

Answer rules: \\
\\

- \textbf{The final answer inside \texttt{<answer>} must be exactly and literally the same as the original assistant answer.} \\
- \textbf{Do not change, rephrase, shorten, expand, normalize, or reinterpret it.} \\
- \textbf{Write only the final answer inside \texttt{<answer>}.} \\
\\

Hard global constraints:\\
\\
- Output nothing before \texttt{<think>} and nothing after \texttt{</answer>}.\\
- Use only the tags: \texttt{<think>}, \texttt{</think>}, \texttt{<FOCUS>}, \texttt{</FOCUS>}, \texttt{<answer>}, \texttt{</answer>}.\\
- The number of \texttt{<FOCUS>} blocks MUST equal:\\
\hspace*{1.5em} - the number of \texttt{<SOT>} regions in the original assistant response, if there is at least one, OR\\
\hspace*{1.5em} - exactly 1, if there are no \texttt{<SOT>} regions.\\
- Never restate these rules or constraints inside the output.\\

\end{tcolorbox}


%
\begin{tcolorbox}[
  breakable,
  title={Prompt For Constructing Dataset From PixelReasoner SFT Dataset Using Single-Pass Distillation},
  colback=gray!10,
  colframe=gray!60,
  fonttitle=\bfseries,
]
You must rewrite the original assistant reasoning and output in exactly the following structure and nothing else: \\
\\
\texttt{<think>} \\
\texttt{...your rewritten reasoning with one or more <FOCUS> blocks inserted in-line...} \\
\texttt{</think>} \\
\\
\texttt{<answer>} \\
\texttt{...final answer...} \\
\texttt{</answer>} \\
\\
Your task: \\
1. Use the original assistant response (with \texttt{<SOT>} markers) as guidance. \\
2. Rewrite the reasoning in your own words. \\
3. Insert \texttt{<FOCUS>} blocks at the correct places. \\
\\
\textbf{STYLE CONSTRAINTS (NO "ZOOM"):} \\
- \textbf{FORBIDDEN WORDS}: Do NOT use terms related to digital tools or image manipulation, such as \textbf{``zoom in", ``crop", ``cut", ``resize", ``scale", ``tool", ``selection"}. \\
- \textbf{REQUIRED STYLE}: Use natural visual attention language. Instead of ``I will zoom in on...", use phrases like: \\
\hspace*{1.5em} - ``I examine..." \\
\hspace*{1.5em} - ``I focus my attention on..." \\
\hspace*{1.5em} - ``Looking closer at..." \\
\hspace*{1.5em} - ``Inspecting the..." \\
\\
\textbf{Answer Rules (NO BOXED FORMAT):} \\
- You must output the final answer text inside \texttt{<answer>...</answer>}. \\
- \textbf{CRITICAL}: If the original answer contains \texttt{\textbackslash\textbackslash boxed\{TEXT\}}, \textbf{REMOVE} the \texttt{\textbackslash\textbackslash boxed\{\}} wrapper and strictly output ONLY the \texttt{TEXT}. \\
- Example: If original is \texttt{\textbackslash\textbackslash boxed\{apple\}}, output \texttt{apple}. If original is \texttt{\textbackslash\textbackslash boxed\{B\}}, output \texttt{B}. \\
- Do NOT include any coordinates or bounding boxes in the answer. \\
\\
\textbf{Standard Rules:} \\
- If the original response contains N region markers, produce N \texttt{<FOCUS>} blocks. \\
- \texttt{<FOCUS>} content must be ONE sentence identifying the object/region (e.g., ``The red boat is the element that must be examined."). \\
- Do NOT include coordinates or OCR text inside \texttt{<FOCUS>}. \\
- Outside \texttt{<FOCUS>}, verify the details seen in that region. \\

\end{tcolorbox}


\begin{tcolorbox}[
  breakable,
  title={ Prompt For Constructing Dataset From PixelReasoner SFT Dataset Using Recrop Distillation},
  colback=gray!10,
  colframe=gray!60,
  fonttitle=\bfseries,
]
You must rewrite the original assistant reasoning and output in exactly the following structure and nothing else: \\
\\
\texttt{<think>} \\
\texttt{...your rewritten reasoning...} \\
\texttt{</think>} \\
\\
\texttt{<answer>} \\
\texttt{...final answer...} \\
\texttt{</answer>} \\
\\
\textbf{CRITICAL INSTRUCTION: CLEANING UP TRIAL-AND-ERROR} \\
The ``Original Guidance" likely contains a trial-and-error process (e.g., look $\to$ fail $\to$ look again $\to$ success). \\
\textbf{Your task is to REWRITE this history as if you got it right on the first try.} \\
1. \textbf{Ignore Failures}: Completely ignore early attempts that failed or saw nothing. \\
2. \textbf{Identify Success}: Find the final \texttt{<SOT>} region that actually provided the answer. \\
3. \textbf{Synthesize}: Write a direct, successful reasoning path. \\
\\
\textbf{STRICT REASONING FLOW (IMPORTANT):} \\
You must strictly follow this logical order in your \texttt{<think>} block: \\
1. \textbf{Context}: Briefly scan the whole image or general area. \\
2. \textbf{Focus (Action)}: Identify the target region. \textbf{Insert the \texttt{<FOCUS>} block HERE.} \\
3. \textbf{Observation (Result)}: \textbf{AFTER} the \texttt{<FOCUS>} block, describe what you see inside that region (e.g., reading text, checking numbers, describing details). \\
4. \textbf{Conclusion}: Derive the final answer. \\
\\
\textbf{STYLE CONSTRAINTS (NO ``ZOOM"):} \\
- \textbf{FORBIDDEN}: Do NOT use digital tool terms like ``zoom in", ``crop", ``cut", ``resize", ``tool". \\
- \textbf{REQUIRED}: Use natural vision terms like ``examine", ``focus attention on", ``inspect", ``observe". \\
\\
\textbf{Answer Rules (NO BOXED):} \\
- You must output the final answer text inside \texttt{<answer>...</answer>}. \\
- \textbf{CRITICAL}: If the original answer contains \texttt{\textbackslash\textbackslash boxed\{TEXT\}}, \textbf{REMOVE} the \texttt{\textbackslash\textbackslash boxed\{\}} wrapper and strictly output ONLY the \texttt{TEXT}. \\
- Example: If original is \texttt{\textbackslash\textbackslash boxed\{A\}}, output \texttt{A}. \\
- Do NOT include any coordinates or bounding boxes in the answer. \\
\end{tcolorbox}
\section{AttnDetector}\label{app:AttnDetector}
AttnDetector takes an attention heatmap as input and predicts the corresponding ROI bounding box. To train AttnDetector, we construct paired heatmap--box supervision. Specifically, we run the trained ConFoAttn on the ConFoAttn dataset, compute the designated-layer attention from the \texttt{<FOCUS>}...\texttt{</FOCUS>} span to image tokens as the heatmap, and use the corresponding target box label provided by the dataset as supervision, forming the AttnDetector dataset. For example, in Fig.~\ref{fig:method}(b), ConFoAttn computes a text-to-image attention map using the \texttt{<FOCUS>}...\texttt{</FOCUS>} span in Fig.~\ref{fig:method}(a) as the query, which serves as the input to AttnDetector (the green patch near the top indicates high attention in Fig.~\ref{fig:method}(b)). The corresponding coordinates are used as supervision for AttnDetector’s box prediction.

We then train AttnDetector as a DETR-style single-box regressor with an $\mathcal{L}_1$ loss and a GIoU loss. Besides the heatmap, we concatenate three auxiliary channels to form a 4-channel input tensor: a padding mask indicating valid (1) versus padded (0) positions, and normalized $x$/$y$ coordinate grids that provide explicit spatial cues for localization. 

The 4-channel input is first encoded by a lightweight convolutional downsampling stem, producing a compact feature map.
Concretely, the stem is implemented as a stack of $n$ $3{\times}3$ convolutional blocks with stride 2, each followed by GroupNorm and a SiLU nonlinearity. We then flatten the feature map into a token sequence and feed it into a lightweight Transformer encoder with two layers of multi-head self-attention (8 heads; hidden size 256).
A single learnable query in a two-layer Transformer decoder attends to the encoded tokens via multi-head cross-attention (8 heads; hidden size 256) and summarizes the ROI into one query embedding. Finally, an MLP head (Linear--SiLU--Linear) outputs four raw logits, which are passed through a sigmoid to obtain normalized box parameters $(\hat{\mathbf{u}}=\hat{c}_x,\hat{c}_y,\hat{w},\hat{h})$ in $(0,1)$.
We convert $\hat{\mathbf{u}}$ to corner coordinates $\hat{\mathbf{b}}=(\hat{x}_1,\hat{y}_1,\hat{x}_2,\hat{y}_2)$ via
\begin{equation}
\begin{aligned}
\hat{x}_1 &= \hat{c}_x-\tfrac{1}{2}\hat{w}, \qquad & \hat{x}_2 &= \hat{c}_x+\tfrac{1}{2}\hat{w},\\
\hat{y}_1 &= \hat{c}_y-\tfrac{1}{2}\hat{h}, \qquad & \hat{y}_2 &= \hat{c}_y+\tfrac{1}{2}\hat{h}.
\end{aligned}
\end{equation}
Let $\mathbf{b}=(x_1,y_1,x_2,y_2)$ the ground truth. We optimize the detector with an $\ell_1$ regression loss and a GIoU loss:
\begin{equation}
\begin{aligned}
\mathcal{L}_{\ell_1} &= \left\lVert \hat{\mathbf{b}}-\mathbf{b}\right\rVert_1, \\
\mathcal{L}_{\mathrm{GIoU}} &= 1-\mathrm{GIoU}(\hat{\mathbf{b}},\mathbf{b}), \\
\mathcal{L}_{\mathrm{det}} &= \mathcal{L}_{\ell_1}+\mathcal{L}_{\mathrm{GIoU}} \, .
\end{aligned}
\end{equation}



\end{document}